\pdfoutput=1
\documentclass[11pt]{article}

\usepackage{acl}

\usepackage{times}
\usepackage{latexsym}

\usepackage[T1]{fontenc}
\usepackage[utf8]{inputenc}

\usepackage{microtype}

\usepackage{pgfplots}
\pgfplotsset{compat=1.14} 
\usepackage{graphicx}
\usepackage{xspace}
\usepackage{xurl}                           % simple URL typesetting
\usepackage{pifont}
\usepackage{amsmath}
\usepackage{amssymb}
\usepackage{textcomp}
\usepackage{multirow}
\usepackage{graphicx}
\usepackage{microtype}
\usepackage{enumitem}
\usepackage{array}
\usepackage{lipsum} 
\usepackage{rotating}
\usepackage{booktabs}
\usepackage{xcolor,colortbl}
\usepackage{graphicx}
\usepackage{lscape}
\definecolor{orange}{HTML}{FFCC99}
\definecolor{green}{HTML}{DBEBC8}
\definecolor{purple}{HTML}{E8D8F3}

\title{An Exploration of Hierarchical Attention Transformers \\ for Efficient Long Document Classification}

\author{
  Ilias Chalkidis\thanks{\hspace{0.5em}Corresponding author: ilias.chalkidis[at]di.ku.dk}$^{\;\;\dagger\;}$ \quad
  Xiang Dai$^{\;\ddagger\;}$ \quad
  Manos Fergadiotis$^{\;\diamond\;}$ \\
  \bf Prodromos Malakasiotis$^{\;\diamond\;}$ \quad
  Desmond Elliott$^{\;\dagger\;\measuredangle\;}$ \\
${\;\dagger}$ Department of Computer Science, University of Copenhagen, Denmark \\
${\;\dagger}$ CSIRO Data61, Sydney, Australia\\
$^{\diamond\;}$ Department of Informatics, Athens University of Economics and Business, Greece\\
$^{\measuredangle\;}$ Pioneer Centre for AI, Copenhagen, Denmark \\
}

\begin{document}
\maketitle

\begin{abstract}
Non-hierarchical sparse attention Transformer-based models, such as Longformer and Big Bird, are popular approaches to working with long documents. There are clear benefits to these approaches compared to the original Transformer in terms of efficiency, but Hierarchical Attention Transformer (HAT) models are a vastly understudied alternative.
We develop and release fully pre-trained HAT models that use segment-wise followed by cross-segment encoders and compare them with Longformer models and partially pre-trained HATs. In several long document downstream classification tasks, our best HAT model outperforms equally-sized Longformer models while using 10-20\% less GPU memory and processing documents 40-45\% faster. In a series of ablation studies, we find that HATs perform best with cross-segment contextualization throughout the model than alternative configurations that implement either early or late cross-segment contextualization. Our code is on GitHub:~\url{https://github.com/coastalcph/hierarchical-transformers}.
\end{abstract}

\section{Introduction}
\label{sec:introduction}

\emph{Long Document Classification} is the classification of a single long document typically in the length of thousands of words, e.g., classification of legal \cite{chalkidis-etal-2022-lexglue} and biomedical documents \cite{johnson-mit-2016-mimic-iii}, or co-processing of long and shorter chunks of texts, e.g., sequential sentence classification~\cite{cohan-etal-2019-pretrained}, document-level multiple-choice QA \cite{pang-2022-quality}, and document-level NLI \cite{koreeda-manning-2021-contractnli-dataset}.

One approach to working with long documents is to simply expand standard Transformer-based language models (BERT of \citet{devlin_bert_2019}, RoBERTa of \citet{liu_roberta_2019}, etc.) but this is problematic for long sequences, given the $O(N^2)$ self-attention operations. To address this computational problem, researchers have introduced efficient Transformer-based architectures. Several sparse attention networks, such as Longformer of \citet{beltagy_longformer_2020}, or BigBird of \citet{zaheer2020bigbird}, have been proposed relying on a combination of different attention patterns (e.g., relying on local (neighbor), global and/or randomly selected tokens). Another approach relies on Hierarchical Attention Transformers (HATs) that use a multi-level attention pattern: segment-wise followed by cross-segment attention.
Ad-hoc (partially pre-trained), and non-standardized variants of HAT have been presented in the literature \cite{chalkidis_neural_2019, wu-etal-2021-hi, chalkidis-etal-2022-lexglue, ernie-sparse, dai-et-al-2022-hierarchical}, but the potential of such models is still vastly understudied.

\begin{figure}[t]
    \centering
    \includegraphics[width=\columnwidth]{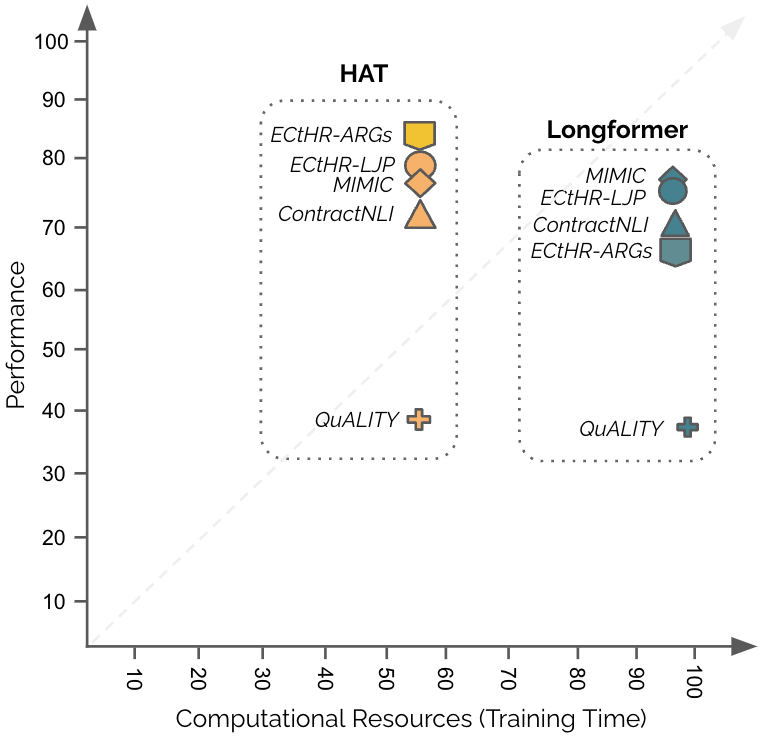}
    \vspace{-7mm}
    \caption{Performance - Efficiency trade-off for HAT and Longformer on downstream tasks.}
    \label{fig:intro}
    \vspace{-6mm}
\end{figure}

In this work, we examine the potential of fully (end-to-end) pre-trained HATs and aim to answer three main questions: (a) Which configurations of segment-wise and cross-segment attention layers in HATs perform best? (b) What is the effect of pre-training HATs end-to-end, compared to ad-hoc (partially pre-trained), i.e., plugging randomly initialized cross-segment transformer blocks during fine-tuning? (c) Are there computational or downstream perfomance benefits of using HATs compared to widely-used sparse attention networks, such as Longformer and BigBird?

\section{Related Work}

\subsection{Sparse Attention Transformers}

\noindent\textbf{Longformer} of \citet{beltagy_longformer_2020} consists of local (window-based) attention and global attention that reduces the computational complexity of the model and thus can be deployed to process up to $4096$ tokens. Local attention is computed in-between a window of neighbour (consecutive) tokens. Global attention relies on the idea of global tokens that are able to attend and be attended by any other token in the sequence. Windowed (local) attention does not leverage hierarchical information in any sense, and can be considered greedy.\vspace{2mm}

\noindent\textbf{BigBird} of \citet{zaheer2020bigbird} is another sparse-attention based Transformer that uses a combination of a local, global and random attention, i.e., all tokens also attend a number of random tokens on top of those in the same neighbourhood. 
Both models are warm-started from the public RoBERTa checkpoint and are further pre-trained on masked language modelling. They have been reported to outperform RoBERTa on a range of tasks that require modelling long sequences.\vspace{2mm}

\begin{figure}[t]
    \centering
    \includegraphics[width=\columnwidth]{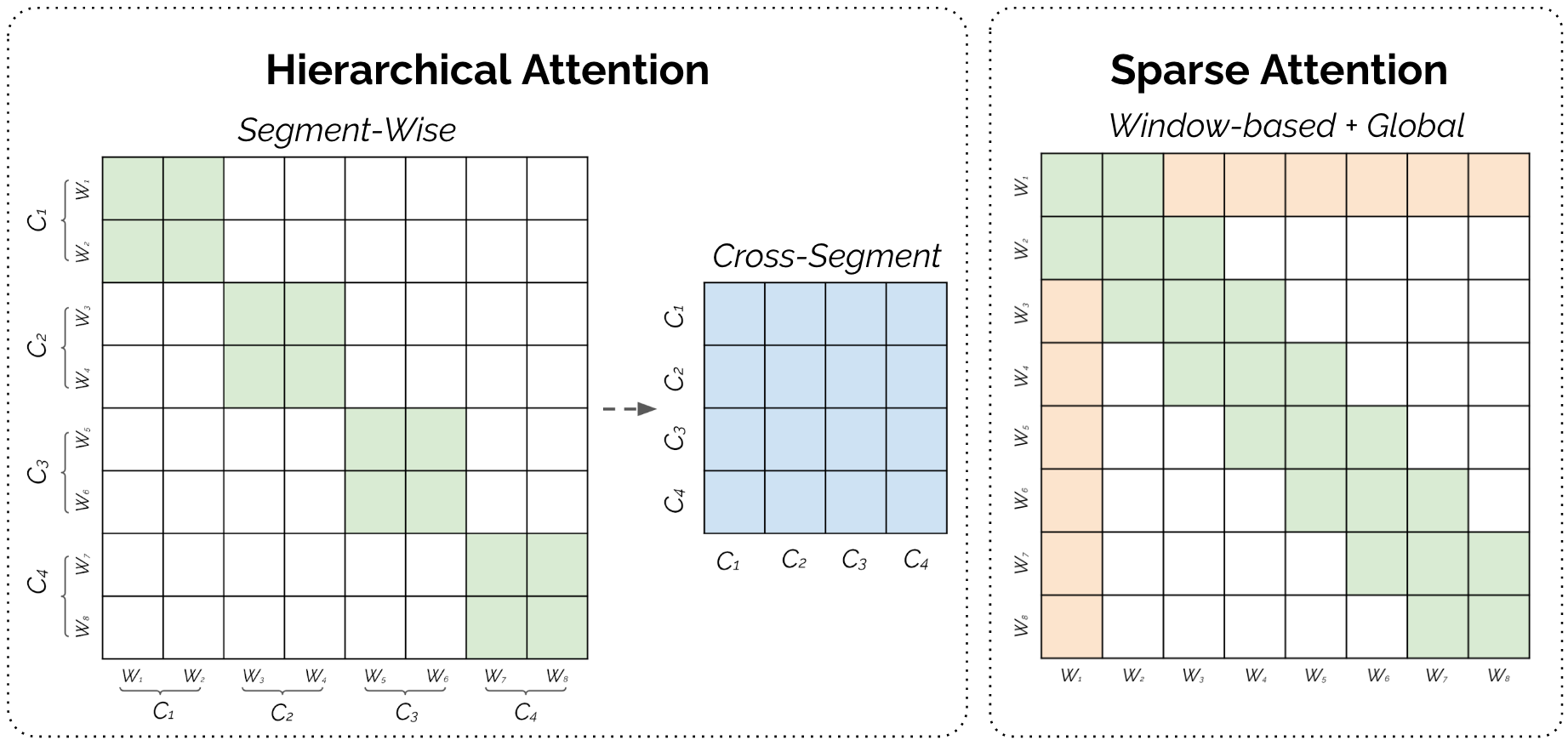}
    \vspace{-7mm}
    \caption{Attention patterns for the examined architectures: \emph{Hierarchical} (Segment-wise followed by cross-segment attention) and \emph{Sparse} (Combination of windowed and global attention) Attention Transformers.}
    \label{fig:archs}
    \vspace{-5mm}
\end{figure}

\noindent In both cases (models), the attention scores for local (neighbor), global, and randomly selected tokens are combined (added), i.e., attention blends only word-level representations (Figure~\ref{fig:intro}). BigBird is even more computationally expensive with borderline improved results in some benchmarks, e.g., LRA of \citet{tay2021long}, but not in others, e.g., LexGLUE of \citet{chalkidis-etal-2022-lexglue}.

\begin{figure*}[ht]
    \centering
    \includegraphics[width=\textwidth]{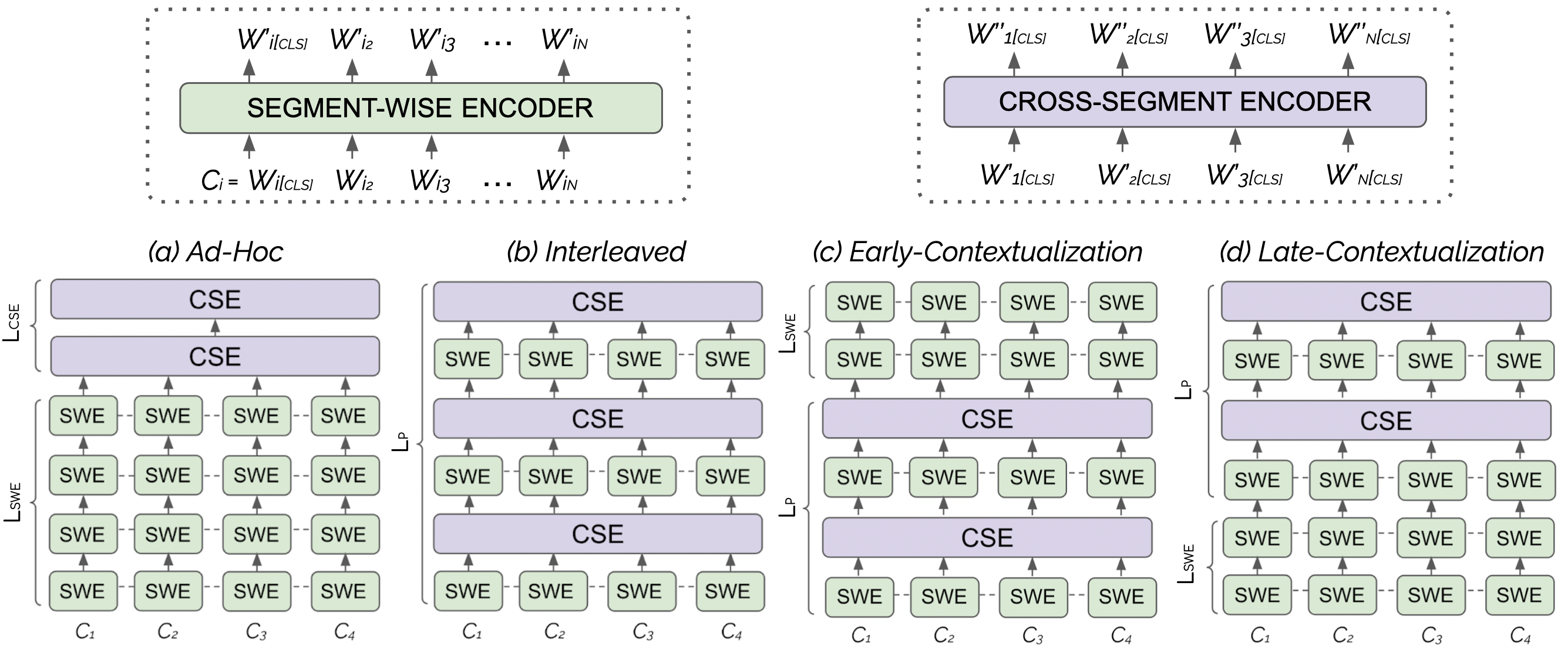}
    \vspace{-7mm}
    \caption{Top: The two main modules (building blocks) of Hierarchical Attention Transformers (HAT): the \emph{Segment-wise} (\colorbox{green}{SWE}), and the \emph{Cross-segment} (\colorbox{purple}{CSE}) encoders. Bottom: The four examined HAT variants.}
    \label{fig:layouts}
    % \vspace{-2mm}
\end{figure*}

\subsection{Hierarchical Attention Transformers}

Hierarchical Attention Transformers (HATs) are directly inspired by Hierarchical Attention Networks (HANs) of \citet{yang-etal-2016-hierarchical}. The main idea is to process (encode) document in a hierarchical fashion, e.g., contextualize word representations per sentence, and then sentence-level representations across sentences. \citet{chalkidis_neural_2019} were probably the first to use HATs as a viable option for processing long documents based on pre-trained Transformer-based language models. They show improved results using a hierarchical variant of BERT compared to BERT (fed with truncated documents) or HANs. Similar models were used in the work of \citet{chalkidis-etal-2022-lexglue}, where they compared hierarchical variants of several pre-trained language models (BERT, RoBERTa, etc.) showcasing comparable results to Longformer and BigBird in long document classification tasks. Recently, \citet{dai-et-al-2022-hierarchical} compared ad-hoc RoBERTa-based HATs with Longformer and reported comparable performance in four document classification tasks. 

\citet{wu-etal-2021-hi} proposed a HAT architecture, named Hi-Transformers, a shallow version of our interleaved variant presented in detail in Section~\ref{sec:layouts}. They showed that their model performs better compared to Longformer and BigBird across three classification tasks. Although their analysis relies on non pre-trained models, i.e., all models considered are randomly initialized and directly fine-tuned on the downstream tasks, thus the impact of pre-training such models is unknown.

\citet{ernie-sparse} propose a similar architecture, named  Hierarchical Sparse Transformer (HST). \citeauthor{ernie-sparse} showed that HST has improved results in the long range arena (LRA) benchmark, text classification and QA compared to Longformer and BigBird. Their analysis considers a single layout (topology) and is mainly limited on datasets where documents are not really long (<1000 tokens). In our work, we consider several HAT layouts (configurations) and evaluate our models in several segment-level, document-level, and multi-segment tasks with larger documents (Table~\ref{tab:datasets}).

\subsection{Other Approaches}

Several other efficient Transformer-based models have been proposed in the literature \cite{katharopoulos_et_al_2020, Kitaev2020Reformer,choromanski2021rethinking}. We refer readers to~\citet{xiong-fb-2021-local-attention,tay-dehghani-efficient-survey} for a survey on efficient attention variants. Recently other non Transformer-based approaches \cite{gu2022efficiently,gupta2022diagonal} have been proposed for efficient long sequence processing relying on structured state spaces \cite{gu2021combining}. In this work, we do not compare with such architectures (Transformer-based or not), since there are no standardized implementations or publicly available pre-trained models to rely on at the moment. 
There are several other Transformer-based encoder-decoder models \cite{guo-etal-2022-longt5, pang-etal-2022} targeting generative tasks, e.g., long document summarization \cite{shen2022multilexsum}, which are out of the scope in this study.

\section{Hierarchical Attention Transformers}

\subsection{Architecture}
\label{sec:hier-architecture}

Hierarchical Attention Transformers (HATs) consider as input a sequence of tokens ($S$), organized in $N$ equally-sized segments (chunks) ($S\!=\![C_1, C_2, C_3,\dots,C_N]$). Each sub-sequence (segment) is a sequence of $K$ tokens ($C_i\!=\![W_{i\texttt{[CLS]}},W_{i1}, W_{i2}, W_{i3},\dots, W_{iK-1}]$), i.e. each segment has its own segment-level representative \texttt{[CLS]} token. A HAT is built using two types of neural modules (blocks): (a) the \emph{Segment-wise encoder (SWE)}: A shared Transformer \cite{vaswani-etal-2017} block processing each segment ($C_i$) independently, and (b) the \emph{Cross-segment encoder (CSE)}: A Transformer block processing (and contextualizing) segment-level representative tokens ($W_{i\texttt{[CLS]}}$). The two components can be used in several different layouts (topologies). We present HAT variants (architectures) in Section~\ref{sec:layouts}.

HATs use two types of absolute positional embeddings to model the position of tokens: \emph{segment-wise} position embeddings ($P^{sw}_{i}\!\in\!\mathrm{R}^{H}, i\in[1,K]$) to model token positioning per segment, and \emph{cross-segment} position embeddings ($P^{cs}_{i}\!\in\!\mathrm{R}^{H}, i\in[1,N]$) to model the position of a segment in the document. $P^{sw}$ embeddings are additive to word ones, like in most other Transformer-based models, such as BERT. Similarly, $P^{cs}$ embeddings are added to the segment representations ($W'_{i\texttt{[CLS]}}$) before they are passed to a CSE, and they are shared across all CSEs of the model. A more detailed depiction of HAT including positional embeddings is presented in Figure~\ref{fig:model_example} of Appendix~\ref{sec:pos_embeds}.

\subsection{Examined Layouts}
\label{sec:layouts}

We first examine several alternative layouts of HAT layers, i.e., the placement of SWE and CSE:\vspace{4mm}

\noindent \textbf{Ad-Hoc (AH)}: An ad-hoc (partially pre-trained) HAT \cite{chalkidis-etal-2022-lexglue} comprises an initial stack of shared $L_{\mathrm{SWE}}$ segment encoders from a pre-trained transformer-based model, followed by $L_{\mathrm{CSE}}$ ad-hoc segment-wise encoders. In this case the model initially encodes and contextualize token representations per segment, and then builds higher-order segment-level representations (Figure~\ref{fig:layouts}(a)).\vspace{2mm}

\noindent \textbf{Interleaved (I)}: An interleaved HAT comprises a stack of $L_{\mathrm{P}}$ paired segment-wise and cross-segment encoders. In this case, contrary to the ad-hoc version of HAT, cross-segment attention (contextualization) is performed across several levels (layers) of the model (Figure~\ref{fig:layouts}(b)).\vspace{2mm}

\noindent \textbf{Early-Contextualization (EC)}: An early-contextualized HAT comprises an initial stack of  $L_{\mathrm{P}}$ paired segment-wise and cross-segment encoders, followed by a stack of $L_{\mathrm{SWE}}$ segment-wise encoders. In this case, cross-segment attention (contextualization) is only performed at the initial layers of the model (Figure~\ref{fig:layouts}(c)).\vspace{2mm}

\noindent \textbf{Late-Contextualization (LC)}: A late-contextualized HAT comprises an initial stack of $L_{\mathrm{SWE}}$ segment-wise encoders, followed by a stack of  $L_{\mathrm{P}}$ paired segment and segment-wise encoders. In this case, cross-segment attention (contextualization) is only performed in the latter layers of the model (Figure~\ref{fig:layouts}(d)).\vspace{2mm}

\noindent We present task-specific HAT architectures (e.g., for token/segment/document classification, and multiple-choice QA tasks) in Appendix~\ref{sec:task-specific-archs}.

\subsection{Tokenization / Segmentation}

Since HATs consider a sequence of segments, we need to define a segmentation strategy, i.e. how to group tokens (sub-words) into segments. Standard approaches consider sentences or paragraphs as segments. We opt for a dynamic segmentation strategy that balances the trade-off between the preservation of the text structure (avoid sentence truncation), and the minimization of padding, which minimizes document truncation as a result. We split each document in $N$ segments by grouping sentences up to $K$ total tokens.\footnote{Any sentence splitter can be used. In our work, we consider the NLTK (\url{https://www.nltk.org/}) English sentence splitter. We present examples in Appendix~\ref{sec:model}.}  Following \citet{dai-et-al-2022-hierarchical}, our models consider segments of $K\!=\!128$ tokens each; such a window was shown to balance the computational complexity with task performance.

\section{Experimental Set Up}

\subsection{Evaluation Tasks}
\label{sec:evaluation_tasks}

We consider three groups of evaluation tasks: (a) \emph{Upstream} (pre-training) tasks, which aim to pre-train (warm-start) the encoder in a generic self-supervized manner; (b) \emph{Midstream} (quality-assessment) tasks, which aim to estimate the quality of the pre-trained models; and (c) \emph{Downstream} tasks, which aim to estimate model's performance in realistic (practical) applications.\vspace{2mm}

\noindent \textbf{Upstream (Pre-training) Task}: We consider \emph{Masked Language Modeling (MLM)}, a well-established bidirectional extension of traditional language modeling proposed by \citet{devlin_bert_2019} for Transformer-based text encoders. Following \citet{devlin_bert_2019}, we mask 15\% of the tokens.\vspace{2mm}

\noindent \textbf{Midstream Tasks}: We consider four alternative mid-stream tasks. These tasks aim to assess the quality of word, segment, and document representations of pre-trained models, i.e., models pre-trained on the MLM task.\footnote{We present additional details (e.g., dataset curation) for the midstream tasks in Appendix~\ref{sec:datasets}.\vspace{2mm}

} 

\begin{table*}[t]
    \centering
    \resizebox{\textwidth}{!}{
    \begin{tabular}{ll|l|c|c|c}
        \toprule
         \multicolumn{2}{c|}{\bf Dataset Name} & \bf Task Type & \bf No of Classes & \bf No of Samples & \bf Avg. Doc. Length \\
         \midrule
         MIMIC-III & \citet{johnson-mit-2016-mimic-iii} & Document Classification & 19 & 30,000/10,000/10,000 & 3,522 \\
         ECtHR-LJP & \citet{chalkidis-etal-2021-paragraph} & Document Classification & 10 & 9,000/1,000/1,000 & 2,104\\
         ContractNLI & \citet{koreeda-manning-2021-contractnli-dataset} & Document NLI & 3 & 7,191/2,091/1,037 & 2,220 \\
         QuALITY & \citet{pang-2022-quality} & Multiple-Choice QA & 4 & 2,523/1,058/1,028 & 6,821 \\
         ECtHR-ARG & \citet{habernal-etal-2022-argument} & Paragraph Classification & 8 & 900/100/100 & 1,285 \\
         \bottomrule
    \end{tabular}
    }
    \vspace{-2mm}
    \caption{Specifications for the examined long document downstream tasks (datasets). We report the task type, number of classes and number of samples across training, development, and test subsets. We also report the average document length measures in BPEs produced by the RoBERTa tokenizer.}
    \label{tab:datasets}
\end{table*}

\begin{itemize}[leftmargin=8pt]
    \item \emph{Segment Masked Language Modeling (MLM)}, an extension of MLM, where a percentage of tokens in a subset (20\%) of segments are masked. We consider two alternatives: 40\% (SMLM-40) and 100\%  (SMLM-100) masking. For this tasks, we predict the identity of the masked tokens. We use cross-entropy loss as the evaluation metric. Intuitively we assess cross-segment contextualization, since we predict masked words of a segment mainly based on the other segments.
    \item \emph{Segment Order Prediction (SOP)}, where the input for a model is a shuffled sequence of segments from a document. The goal of the task is to predict the correct position (order) of the segments, as it was in the original document. For this task, we predict the position per segment as a regression task; hence our evaluation metric is mean absolute error (mae). Intuitively we assess cross-segment contextualization and the quality of segment-level representations since segment order has to resolved given segment relations.
    \item  \emph{Multiple-Choice Masked Segment Prediction (MC-MSP)}, where the input for a model is a sequence of segments from a document with one segment being masked at a time, and a list of five alternative segments (choices) including the masked one. The goal on this task for the model, is to identify the correct segment; the one masked from the original document. For this task, we predict the id of the correct pair (<masked document, choice>) across all pairs; hence our evaluation metric is accuracy. Similarly with SOP we assess cross-segment contextualization and the quality of segment-level representations, since predicting the correct segment has to be resolved based on both document-level semantics and those of the neighbor segments to the masked one.
    \item  \emph{Document Topic Classification (DTC)}, where the input for a model is a full document. The goal on this task for the model is to identify the correct label out of $N$ alternative labels (topics). Intuitively we assess document-level representations, since the relevant topics are inferred by the document-level (pooled) representations. This a single-label multi-class classification task, and the evaluation metric is micro-averaged F1 (F1).
\end{itemize}

\noindent \textbf{Downstream Tasks}: We consider four downstream long classification tasks, covering four task types across three different application domains.\footnote{We present statistics (e.g., number of samples) and other additional details on Appendix~\ref{sec:datasets}.}

\begin{itemize}[leftmargin=8pt]
    \item \emph{MIMIC-III} \cite{johnson-mit-2016-mimic-iii} contains approx.\ 50k discharge summaries from US hospitals. Each summary is annotated with one or more codes (labels) from the ICD-9 taxonomy.
    % \footnote{\url{www.who.int/classifications/icd/en/}} 
    % The International Classification of Diseases, Ninth Revision (ICD-9)  is the official system of assigning codes to diagnoses and procedures associated with hospital utilization in the United States and is maintained by the World Health Organization (WHO). 
    The input of the model is a discharge summary, and the output is the set of the relevant 1st level ICD-9 (19 in total) codes.
    \item \emph{ECtHR-LJP} \cite{chalkidis-etal-2021-paragraph} contains approx.\ 11K cases from the European Court of Human Rights (ECtHR) public database.
    % \footnote{\url{https://hudoc.echr.coe.int/eng}} 
    For each case, the dataset provides a list of \emph{factual} paragraphs (facts) from the case description. Each case is mapped to articles of ECHR that were \emph{allegedly} violated (considered by the court). The input of the model is the list of facts of a case, and the output is the set of \emph{allegedly} violated articles.
    \item \emph{ContractNLI} \cite{koreeda-manning-2021-contractnli-dataset} is a dataset for contract-based Natural Language Inference (NLI). The dataset consists of 607 contracts, specifically Non-Disclosure Agreements (NDAs).
    % , which -similarly to LEDGAR, have been obtained from SEC filings. 
    Each document has been paired with 17 templated \emph{hypothesis} and labeled with one out of three classes (\emph{entailment}, \emph{contradiction}, or \emph{neutral}).  This is a single-label multi-class classification task. The inputs to a model is the full document, and a hypothesis, and the output is the correct out of the three plausible classes.
    \item \emph{QuALITY} \cite{pang-2022-quality} contains approx.~5k questions based on reference documents (books or articles). Each questions is paired with 4 alternative answers, one of which is the correct one. The input of the model is the document (context), the questions, and the four alternative answers, and the output is the the id of the correct answer.
    \item \emph{ECtHR-ARG} \cite{habernal-etal-2022-argument} contains approx.\ 300 cases from the European Court of Human Rights (ECtHR). For each case, the dataset provides a list of \emph{argumentative} paragraphs from the case analysis. Spans in each paragraph has been labeled with one or more out of 13 argument types. We re-formulate this task, as a sequential paragraph classification task, where each paragraph is labelled with one or more labels. The input of the model is the list of paragraphs of a case, and the output is the set of relevant argument types per paragraph.\footnote{We consider the 8 most frequent argument types, since the rest are extremely rare (less than 50 labeled paragraphs).}
    % \item \emph{AIDA} \cite{cao2021autoregressive} contains 388 articles from the Reuters Corpus annotated with approx.~27.4k mentions of named entities. The inputs to a model is a document (incl. a mention) and the descriptors of two candidate entities, and the output is the id of the correct (mentioned) entity. 
\end{itemize}

\begin{table*}[t]
    \centering
    \resizebox{\textwidth}{!}{
    \begin{tabular}{lc|c|c|c|c|c|c|c|c|c|c|c|c|c|c|c|c}
         \multicolumn{2}{c|}{\bf Model Type} & \bf SWE & \bf Params &  \multicolumn{12}{c|}{\bf Layout (Encoder Type per Layer)} & \bf SpeedUp & \bf MemSave \\
         \midrule
         \multicolumn{18}{c}{Baselines} \\
         \midrule
         Longformer &  & 6 & 14.4M & \multicolumn{2}{c|}{\cellcolor{orange} W+G} & \multicolumn{2}{c|}{\cellcolor{orange} W+G} & \multicolumn{2}{c|}{\cellcolor{orange} W+G} & \multicolumn{2}{c|}{\cellcolor{orange} W+G} & \multicolumn{2}{c|}{\cellcolor{orange} W+G} & \multicolumn{2}{c|}{\cellcolor{orange} W+G} & - & - \\
         MiniHAT & AH1 & 6 & 17.7M & \cellcolor{green} SW & \cellcolor{green} SW & \cellcolor{green} SW & \cellcolor{green} SW & \cellcolor{green} SW & \cellcolor{green} SW & \cellcolor{purple} CS & \cellcolor{purple} CS & \cellcolor{purple} CS & \cellcolor{purple} CS & \cellcolor{purple} CS & \cellcolor{purple} CS & 20\% & 2\%  \\
         MiniHAT & AH2 & 8 & >> & \cellcolor{green} SW & \cellcolor{green} SW & \cellcolor{green} SW & \cellcolor{green} SW & \cellcolor{green} SW & \cellcolor{green} SW & \cellcolor{green} SW & \cellcolor{green} SW & \cellcolor{purple} CS & \cellcolor{purple} CS & \cellcolor{purple} CS & \cellcolor{purple} CS & 20\% & -4\%  \\
         \midrule
         \multicolumn{18}{c}{Interleaved} \\
         \midrule
         MiniHAT & I1 & 6 & >> & \cellcolor{green} SW & \cellcolor{purple} CS & \cellcolor{green} SW & \cellcolor{purple} CS & \cellcolor{green} SW & \cellcolor{purple} CS & \cellcolor{green} SW & \cellcolor{purple} CS & \cellcolor{green} SW & \cellcolor{purple} CS & \cellcolor{green} SW & \cellcolor{purple} CS & 20\% & 2\%  \\
         MiniHAT & I2 & 6 & >> & \cellcolor{green} SW & \cellcolor{green} SW & \cellcolor{purple} CS & \cellcolor{purple} CS & \cellcolor{green} SW & \cellcolor{green} SW & \cellcolor{purple} CS & \cellcolor{purple} CS & \cellcolor{green} SW & \cellcolor{green} SW & \cellcolor{purple} CS & \cellcolor{purple} CS & 20\% & 2\%  \\
         MiniHAT & I3 & 8 & >> & \cellcolor{green} SW & \cellcolor{green} SW & \cellcolor{purple} CS & \cellcolor{green} SW & \cellcolor{green} SW & \cellcolor{purple} CS & \cellcolor{green} SW & \cellcolor{green} SW & \cellcolor{purple} CS & \cellcolor{green} SW & \cellcolor{green} SW & \cellcolor{purple} CS & 20\% & -4\%  \\
         MiniHAT & I4 & 9 & >> & \cellcolor{green} SW & \cellcolor{green} SW & \cellcolor{green} SW & \cellcolor{purple} CS & \cellcolor{green} SW & \cellcolor{green} SW & \cellcolor{green} SW & \cellcolor{purple} CS & \cellcolor{green} SW & \cellcolor{green} SW & \cellcolor{green} SW & \cellcolor{purple} CS & 20\% & -6\%  \\
         \midrule
          \multicolumn{18}{c}{Early-Fusion} \\
          \midrule
          MiniHAT & EC1 & 9 & >> & \cellcolor{green} SW & \cellcolor{purple} CS & \cellcolor{green} SW & \cellcolor{purple} CS & \cellcolor{green} SW & \cellcolor{purple} CS & \cellcolor{green} SW & \cellcolor{green} SW & \cellcolor{green} SW & \cellcolor{green} SW & \cellcolor{green} SW & \cellcolor{green} SW & 20\% & -6\%  \\
          MiniHAT & EC2 & 8 & >> & \cellcolor{green} SW & \cellcolor{green} SW & \cellcolor{purple} CS & \cellcolor{purple} CS & \cellcolor{green} SW & \cellcolor{green} SW & \cellcolor{purple} CS & \cellcolor{purple} CS & \cellcolor{green} SW & \cellcolor{green} SW & \cellcolor{green} SW & \cellcolor{green} SW & 20\% & -4\%  \\
          \midrule
         \multicolumn{18}{c}{Late-Fusion} \\
         \midrule
         MiniHAT & LC1 & 9 & >> & \cellcolor{green} SW & \cellcolor{green} SW & \cellcolor{green} SW & \cellcolor{green} SW & \cellcolor{green} SW & \cellcolor{green} SW & \cellcolor{green} SW & \cellcolor{purple} CS & \cellcolor{green} SW & \cellcolor{purple} CS & \cellcolor{green} SW & \cellcolor{purple} CS & 20\% & -6\%  \\
         MiniHAT & LC2 & 8 & >> & \cellcolor{green} SW & \cellcolor{green} SW & \cellcolor{green} SW & \cellcolor{green} SW & \cellcolor{green} SW & \cellcolor{green} SW & \cellcolor{purple} CS & \cellcolor{purple} CS & \cellcolor{green} SW & \cellcolor{green} SW & \cellcolor{purple} CS & \cellcolor{purple} CS & 20\% & -4\%  \\
         \bottomrule
    \end{tabular}
    }
    \vspace{-2mm}
    \caption{Layouts examined for miniature models. \textbf{SWE}: number of segment-wise encoders. \textbf{Layout}: the organization of segment-wise (SW) and cross-segment (CS) encoders. In case of Longformer, there are encoders with paired window-based and global (W+G) attention. \textbf{SpeedUp} is the time improvement (batch/sec), and \textbf{MemSave} is the decrease in memory over Longformer (LF) for masked language modelling using 1$\times$A100 40GB.}
    \label{tab:layouts}
    % \vspace{-2mm}
\end{table*}

\vspace{-1mm}
\section{Experiments}

\subsection{Miniature Language Models (MiniHATs)}
\label{sec:minilms}

We start by conducting a controlled study in which we pre-train different miniature Hierarchical Transformer-based models in a standard MLM setting. We call them \emph{MiniHATs}, in short. The MiniHATs have 12 Transformer blocks (layers) in total, each having 256 hidden units with 4 attention heads. We examine 8 alternative model layouts (4 interleaved, 2 early-contextualization, 
and 2 late-contextualization; see Section \ref{sec:layouts} for additional details), whose exact layout is presented in Table~\ref{tab:layouts}.\vspace{2mm} 

\noindent\textbf{Warm-Start:} Following \citet{beltagy_longformer_2020} and \citet{zaheer2020bigbird}, we warm-start the MiniHATs from pre-trained checkpoints. In preliminary experiments we find that the best warm-up strategy for our models is to warm-start all embedding layers (word, position, type), and all transformer blocks in pairs, i.e., the weights of each original Transformer block are copied to a SWE encoder, and the following CSE encoder, if any.\footnote{The results of the preliminary experiments are presented in Appendix~\ref{sec:experimental_details}.} For warm-starting MiniHATs, we use the miniature BERT models of \citet{turc-etal-2019}. We consider models based on the numbers of SWE encoders, i.e., the I1 variant of MiniHAT with 6 SWE encoders is warm-started from the 6-layer BERT model of \citeauthor{turc-etal-2019}. We train models with sequences up to 1024 tokens (8$\times$ segments of 128 tokens). Similarly to \citeauthor{turc-etal-2019}, we use English Wikipedia (2021 dump) to build the datasets for the MLM, MSLM-40/100, SOP and MC-MSP mid-stream tasks (Section~\ref{sec:evaluation_tasks}).\vspace{2mm}

\noindent\textbf{Baselines:} We also pre-train a 6-layer Longformer model, which is more computationally intensive (almost equal in terms of memory, but 20\% slower) than our 12-layer MiniHATs (Table~\ref{tab:layouts}). For Longformer, we use a similar tokenization strategy, concatenating segments with the special separator token (\texttt{[SEP]}). We use a window size equal to the segment size ($K\!=\!128$ tokens). The \texttt{[CLS]} and all \texttt{[SEP]} tokens are considered as global tokens across all tasks to improve global information flow. 

We also compare with two ad-hoc (AH) HAT models (with no CS encoders for (S)MLM; because CS encoders do not update word-level representations). Since all models are warm-started, we continue pre-training for 50k steps with batches of 128 samples, similar to \citet{beltagy_longformer_2020}.\vspace{2mm}

\begin{table}[t]
    \centering
    \resizebox{\columnwidth}{!}{
    \begin{tabular}{l|c|c|c}
    \toprule
         \bf Model Name & \bf SWE & \bf Train MLM & \bf Dev MLM \\
         \midrule
         Longformer & n/a & 2.44 & 2.21 \\
         %  LF1 & 6 & 2.47 & 2.24 \\
         \midrule
         MiniHAT - AH1 & 6 & 2.41 & 2.18 \\
         MiniHAT - AH2 & 8 & \underline{2.31} & \underline{2.09} \\
         \midrule
         MiniHAT  - I1 & 6 & 2.40 & 2.17 \\ 
         MiniHAT - I2 & 6 & 2.67 & 2.30 \\ 
         MiniHAT - I3 & 8 & \underline{\bf 2.30} & \underline{\bf 2.08} \\ 
         MiniHAT - I4 & 9 & 2.34 & 2.09 \\
         \midrule
         MiniHAT - EC1 & 9 & 2.34 & 2.09 \\ 
         MiniHAT - EC2 & 8 & \underline{2.33} & \underline{2.09} \\ 
         \midrule
         MiniHAT - LC1 & 9 & \underline{2.35} & \underline{2.10} \\ 
         MiniHAT - LC2 & 8 & 2.35 & 2.12 \\
         \bottomrule
    \end{tabular}
    }
    \vspace{-2mm}
    \caption{MLM results of all examined miniature Transformer-based models. We report the training and development MLM losses (cross-entropy). SWE is the number of segment-wise encoders per model.}
    \label{tab:minilm-mlm}
    \vspace{-4mm}
\end{table}

\begin{table*}[t]
    \centering
    \resizebox{0.9\textwidth}{!}{
    \begin{tabular}{l|c|ccc|ccc}
    \toprule
         \multirow{2}{*}{\bf Model Name} & \multirow{2}{*}{\bf SWE} & \bf MLM & \bf SMLM-40 & \bf SMLM-100 & \bf SOP & \bf MC-MSP & \bf DTC \\
         & & loss ($\downarrow$) & loss ($\downarrow$) & loss  ($\downarrow$) & mae  ($\downarrow$) & acc.  ($\uparrow$) & F1 ($\uparrow$) \\
         \midrule
         Longformer & n/a & 2.21 & 4.05 & 6.87 & 0.98 & 87.9 & 76.3 \\
         \midrule
         MiniHAT (AH2) & 8 & 2.09 & 4.08 & 7.08 & 0.89 & 49.1 & 71.8 \\
         MiniHAT (I1) & 6 & 2.17 & 4.09 & \bf 6.38 & 0.89 & 87.1 & 77.1 \\ 
         MiniHAT (I3) & 8 & \bf 2.08 & \bf 4.03 & 6.45 & \bf 0.84 & \bf 89.6 & 77.1 \\ 
         MiniHAT (EC2) & 8 & 2.09 & 4.05 & 6.54 & 0.88 & 79.2 &  76.6 \\
         MiniHAT (LC1) & 9 & 2.10 & 4.07 & 6.68 & 0.90 & 84.2 &  \bf 77.6 \\
         \bottomrule
    \end{tabular}
    }
    \vspace{-2mm}
    \caption{Development results of all examined miniature HAT models on midstream tasks.}
    \label{tab:minilm-results}
    \vspace{-2mm}
\end{table*}

\noindent\textbf{Results:} We present experimental results for the upstream and midstream tasks:
\begin{itemize}[leftmargin=8pt]
\item\emph{Upstream Task (MLM):} In Table~\ref{tab:minilm-mlm}, we present the results for the MLM task. We observe that the general trends suggests that more segment-wise (SW) encoders in favor of cross-segment (CS) encoders are preferable. This is further highlighted given the results of the ad-hoc MiniHATs (AH1 and AH2), which make clear that the benefits of segment-wise contextualization on masked language modelling (in the standard setting) are minimal (the MLM losses are improved by 0.01 comparing AH1 vs I1 and AH2 vs I3). We also observe that interleaved (I) attention patterns (layouts) with 2/1 or 3/1 SW/CS encoders ratio  (cf. I3 and I4) have the best results.

\item\emph{Midstream Tasks:} Based on the initial MLM experiments, we consider and evaluate the following models for all (five) mid-stream tasks: from our baselines, the mini Longformer, the MiniHAT-AH2 -including the four randomly initialized CSE encoders-, and the I1, I3, EC2, LC1 variants of MiniHATs; to cover the best model per layout (AH, I, EC, LC), including the most computationally efficient one (I1).

Table~\ref{tab:minilm-results} shows the development results across all tasks. We observe that in both SMLM tasks, as the percentage of masked tokens increases from 40-100\%, the performance of AH2 is much worse than alternative MiniHATs. These results can be explained as I1, I3, and LC1 have cross-segment encoders; hence they can leverage cross-segment information to compensate for the decrease of unmasked segment-wise context (neighbor tokens).

Moving to the rest of the mid-stream tasks, we observe that the I3 variant has the best overall results, followed by I1 and LC1. In case of the Multiple-Choice Masked Sentence Prediction 
(MC-MSP), and Document Topic Classification (DC), the AH2 models are substantially outperformed by the rest of the models. This result signifies that fully (end-to-end) pre-training of MiniHATs is beneficial, compared to plugging randomly initialized cross-segment (CS) encoders in an ad-hoc fashion; similarly to what \citet{chalkidis-etal-2022-lexglue}, and \citet{dai-et-al-2022-hierarchical} did in their work. 
We also observe that the EC2 variant is substantially outperformed by the rest (I1, I3, LC1) that use throughout (interleaved) or late-contextualization, which sounds reasonable since multiple-choice QA-like tasks heavily rely on cross-segment contextualization, and possibly early-contextualization of poor (early) segment representations is not ideal. 
Lastly, LC1 seems to perform better to DTC compared to I models, while EC2 has the worst performance, which can be explained as early-contextualization may be insufficient for document classification tasks. 
\end{itemize}

\vspace{-1mm}
\noindent\textbf{Primary Observations:} Based on the results with miniature models, we make the following observations: (a) End-to-end pre-training HATs is beneficial compared to ad-hoc solutions; (b) Layouts with more segment-wise encoders perform better than more cross-segment encoders; (c) Interleaving segment-wise and cross-segment blocks is the most promising layout given the overall results; and (d) Interleaved HATs perform better compared to an equally memory-intensive, but slower, Longformer.

\subsection{Larger Language Models}
\label{sec:large_lms}

To further solidify our findings, we extend our work to larger models. We consider the best variant of HAT (I3), given the overall results in Section~\ref{sec:minilms}. Specifically, we train 16-layer models, consisting of 12 segment-wise and 4 cross-segment encoders, in a 4$\times$(3SWE-1CSE) topology), warm-started from the 12-layer RoBERTa model of \citet{liu_roberta_2019}. We also consider a 12-layer Longformer, and a 16-layer ad-hoc HAT, as baselines. 

In this stage, we focus on even larger sequences (up to 4096 tokens; 32$\times$ segments of 128 tokens each). We use C4 \cite{2020t5} to build datasets for upstream and midstream tasks to cover more diverse (and challenging) corpora, similar to those used by \citeauthor{liu_roberta_2019}. As in our experiments in Section~\ref{sec:minilms}, we pre-train models for 50k steps with sequences of at least 1024 tokens.

For reference, we also report results on downstream tasks with the original Longformer of \citet{beltagy_longformer_2020} and BigBird of \citet{zaheer2020bigbird} with the default larger attention window size (512).\vspace{2mm}

\begin{table*}[t]
    \centering
    \resizebox{\textwidth}{!}{
    \begin{tabular}{l|c|ccccc}
         \toprule
         \multirow{3}{*}{\bf Model Name} & \multirow{3}{*}{\bf WS}  & \multicolumn{5}{c}{\emph{Downstream Tasks}} \\
         & & \bf MIMIC & \bf ContractNLI  & \bf ECtHR-LJP & \bf  ECtHR-ARG & \bf  QuALITY \\ 
        %  & \bf  AIDA  \\
        & & F1 ($\uparrow$) & acc. ($\uparrow$)  & F1 ($\uparrow$) & acc. ($\uparrow$) & F1 ($\uparrow$) \\ 
         \midrule
         Longformer (ours) & 128  & 78.9 / 78.7 & \textbf{73.6} / 70.1 & 80.1 / 78.6 & 66.6 / 66.7 & \textbf{36.0} / 38.8  \\
         \midrule
         \textit{Ad-hoc} HAT (ours) & 128  & \textbf{79.0} / 78.8 & 72.0 / 71.3 & 80.2 / \textbf{80.4} & 84.4 / 81.7  & 27.8 / 25.1 \\
         HAT  (ours)  & 128  & \textbf{79.0} / \textbf{78.9} & 72.2 / \textbf{72.1} & \textbf{80.8} / 79.8 & \textbf{84.6} / \textbf{82.6} & 35.8 / \textbf{39.2} \\
         \bottomrule
         Longformer (\citeyear{beltagy_longformer_2020}) & 512 & 78.9 / \bf{78.9} & 71.9 / 71.4 &	80.2 / 78.9 & 80.3 / 80.4 	& tba$\ast$ \\ 
         BigBird  (\citeyear{zaheer2020bigbird})  & 512 & 73.8 / 73.6 & 72.1 / 69.8	& 80.1 / 78.8 & \textbf{84.6} / 81.4 & tba$\ast$ \\
         \bottomrule
    \end{tabular}
    }
    \caption{Results on downstream tasks for all examined RoBERTa-based models. WS refers to the local attention window size. We report both development and test scores (development / test). $\ast$ Results to be announced.}
    \vspace{-2mm}
    \label{tab:roberta-results}
\end{table*}

\begin{table}[t]
    \centering
    \resizebox{\columnwidth}{!}{
    \begin{tabular}{l|ccc}
         \toprule
         \multirow{3}{*}{\bf Model Name}  & \multicolumn{3}{c}{\emph{Upstream/Midstream Tasks}}  \\
         &  \bf MLM  & \bf SOP & \bf MC-MSP \\ 
        & loss  ($\downarrow$)  & mae  ($\downarrow$) & acc. ($\uparrow$)  \\ 
         \midrule
         Longformer & \textbf{1.47} & 4.88 & 99.9 \\
         \textit{Ad-hoc} HAT  & 1.96  & 4.40 & 99.9 \\
         HAT   &  1.54  & \bf 4.35 & 99.9  \\
         \bottomrule
    \end{tabular}
    }
    \caption{Development results on midstream tasks for all examined RoBERTa-based models.}
    \label{tab:roberta-midstream}
    \vspace{-3mm}
\end{table}

\noindent\textbf{Results:} We present experimental results for the upstream, a selection of midstream (SOP, MC-MSP), and downstream tasks:
\begin{itemize}[leftmargin=8pt]

\item\emph{Upstream Task (MLM):} In Table~\ref{tab:roberta-midstream}, we present the results for the MLM pre-training task. We observe that the 12-layer Longformer is slightly better than HAT (approx.\ 0.07 loss reduction). The small improvement can be explained by recalling that Longformer has direct cross-segment contextualization via the window-based local attention, while it also utilized global attention across \textit{all} 12 layers, compared to HAT with 4 cross-segment encoders. Both models perform substantially better than the ad-hoc HAT baseline, which does not consider cross-segment contextualization (approx.\ 0.45 loss reduction).

\item\emph{Midstream Tasks:} We again consider SOP and MC-MSP tasks, which heavily rely on segment-level representations, compared to token-level tasks (SMLM). We omit experiments with DTC since we have several downstream document classification tasks (MIMIC, ContractNLI, ECtHR-LJP). We observe that HAT models (ad-hoc or not) are better in SOP compared to Longformer; similar to our findings in Section~\ref{sec:minilms}. This highlights the benefits of having ``clear'' (independent) segment representations for segment-level tasks. On the second task (MC-MSP), all models in this larger setting make almost perfect predictions, so there is no room for observations.

\item\emph{Downstream Tasks:}  In Table~\ref{tab:roberta-results}, we present the results for all down-stream tasks (Section~\ref{sec:evaluation_tasks}). We observe that there is not a single model that uniformly outperforms the rest of the models. Nonetheless, HAT seems to perform overall better across tasks (document and paragraph classification, NLI, and multiple-choice QA). HATs also outperform the original Longformer of \citet{beltagy_longformer_2020} and BigBird of \citet{zaheer2020bigbird} that use much larger local attention windows (512 tokens), and hence are substantially more computationally intensive.

Both HAT models (ad-hoc or not) severely outperform our Longformer (with similarly sized windows) on ECtHR-ARG (approx.~15\%). We believe this is due to the more standarized processing of the input by HATs, i.e., encoding paragraphs as separate segments. Contrary, the difference is much lower (approx.~1-2\%) compared to the original Longformer of \citet{beltagy_longformer_2020} and BigBird of \citet{zaheer2020bigbird}, since these models use much larger windows, hence the task resembles windowed sentence classification. 

These two observations highlight the importance of cross-segment contextualization in sequential sentence/paragraph classification tasks. Interestigly, our Longformer has comparable results to the more memory-intensive Longformer, and BigBird in all but one tasks, which highlights how we can balance the trade-off of shorter local attention windows using additional global tokens, in a similar fashion with HATs.

\begin{table*}[t]
    \centering
    \resizebox{\textwidth}{!}{
    \begin{tabular}{l|c|c|cc|cc|cc|cc}
         \toprule
         \multirow{3}{*}{\bf Model Name}  & \multirow{3}{*}{\bf WS} & \multirow{3}{*}{\bf Params} & \multicolumn{8}{c}{\emph{Computational Considerations}} \\
         & &  & \multicolumn{2}{c|}{\bf MLM} & \multicolumn{2}{c|}{\bf Doc CLS} & \multicolumn{2}{c|}{\bf Par CLS} & \multicolumn{2}{c}{\bf MCQA} \\
         & & & Mem & Speed & Mem & Speed & Mem & Speed & Mem & Speed \\
         \midrule
         Longformer (ours) & 128 & 148M & - & - & - & -  & - & -  & - & - \\
         \textit{Ad-hoc} HAT (ours)  & 128 & 152M & +10\% & +39\% & +17\% & +43\%  & +18\% & +43\% & +20\% & +46\% \\
         HAT (ours) & 128 & >> & >> & >>  & >> & >> & >> & >>  & >> & >> \\
         \midrule
         Longformer (\citeyear{beltagy_longformer_2020}) & 512 & 148M & \multicolumn{2}{c|}{---}  & -66\% & -305\%   & -70\%  & -87\%   & \multicolumn{2}{c}{tba$\ast$} \\ 
         BigBird  (\citeyear{zaheer2020bigbird})   & 512 & 128M & \multicolumn{2}{c|}{---}  & -76\%  & -276\%   & -75\% &  -73\%   & \multicolumn{2}{c}{tba$\ast$} \\
         \bottomrule
    \end{tabular}
    }
\caption{Computational considerations (number of parameters, and memory and speed improvements over our Longformer) for RoBERTa-based models. WS refers to the local attention window size. $\ast$ Results to be announced.}
    \label{tab:roberta-computational}
    \vspace{-4mm}
\end{table*}

Multiple-choice QA is the only task type where we observe a substantial difference between fully pre-trained models (Longformer, HAT) and the partially pre-trained one (\textit{ad-hoc} HAT). We hypothesize that there are two main reasons: (a) the model follows a stacked layout, where contextualization across document segments (cross-segment attention), the query and the answer choice is performed in the latter stage of the model; (b) the cross-segment encoders are not pre-trained; hence the model ``learns'' how to perform cross-segment contextualization during fine-tuning, which may be particularly important in tasks that heavily rely on cross-segment contextualization (e.g., in multiple-choice QA, where the model has to consider the relative importance of document, query and alternative option).
\end{itemize}

\noindent\textbf{Computational Considerations:} HATs give comparable or better performance than Longformer in downstream tasks. We now consider whether there are computational benefits to HATs based on the statistics presented in Table~\ref{tab:roberta-computational} (top part):

\begin{itemize}[leftmargin=8pt]

\item\emph{Upstream Task (MLM):} In terms of efficiency, HATs use approx.\ 10\% less memory (e.g., 1GB per 10GBs of VRAM), and are approx.\ 40\% faster compared to Longformer in these experiments with larger models. In other words we have a model with comparable performance in the pre-training task, which is much more efficient, especially speed-wise. Given these computational considerations, one could possibly train HATs for almost twice as many steps as Longformer with a similar computational budget (GPU hours), and possibly get much better results.

\item\emph{Downstream Tasks:} With respect to fine-tuning across downstream tasks, HATs use approx.\ 20\% less memory (e.g., 2GB per 10GBs of VRAM), and are approx.\ 45\% faster compared to our Longformer, i.e., one can train HAT models almost twice as fast, given less compute.
\end{itemize}
\vspace{-2mm}
Moving to model deployment (inference), we find that HATs use 10-20\% less memory, and are 20-30\% faster.\footnote{Detailed results presented in Appendix~\ref{sec:compstats}.} In other words, even after the training phase, there are substantial computational gains for deploying HATs over our Longformer.\vspace{2mm}
% Recall that BigBird models, not considered in this study, are even more memory intensive.

\noindent Comparing with the original Longformer of \citet{beltagy_longformer_2020} and BigBird of \citet{zaheer2020bigbird} we observe even greater gains. These models are far more computationally expensive even compared to our Longformer, since they use much larger windows (512 tokens, 4$\times$ larger compared to ours).
Overall, in terms of computational consideration HATs are superior to Longformer and its variants (e.g., BigBird). This has a real-life impact with economical, environmental, and other implications (e.g., access to technology, etc.). \vspace{2mm}

\noindent\textbf{Release of Resources:} Our implementation of HATs relies on the HuggingFace Transformers \cite{wolf_transformers_2020} library; we release our code for reproducibility.\footnote{\url{https://github.com/coastalcph/hierarchical-transformers}}  All examined language models are available on HuggingFace Hub.\footnote{\url{https://huggingface.co/kiddothe2b}}

\section{Conclusions}

In this work, we examined Hierarchical Attention Transformers (HATs) in terms of efficacy (performance) and efficiency (computational considerations) comparing to Longformer, a widely-used sparse attention Transformer. We now conclude answering the three main questions related to the development and potential of such models:
% \vspace{-2mm}
\begin{itemize}[leftmargin=8pt]
\item \emph{(a) Which configurations of segment- and cross-segment attention layers in HATs perform best?} We find that HAT models with cross-segment contextualization throughout the model performs best comparing to other variants (Section~\ref{sec:minilms}).
\item \emph{(b) What is the effect of pre-training HATs end-to-end, compared to ad-hoc (partially pre-trained) ones?} We find that pre-trained HAT models perform substantially better in our small-scale study (Section~\ref{sec:minilms}). The results on larger models are more comparable in most downstream tasks, except document multiple-choice QA on QuALITY.
\item\emph{(c) Are there computational or downstream benefits of using HATs compared to Longformer?} We find that our best pre-trained HAT model performs comparably or better than an equally-sized Longformer across several downstream long document classification tasks, while being substantially faster (40-45\% time decrease) and less memory intensive (10-20\% less GPU memory).
\end{itemize}

\section*{Limitations}

In this work, we consider MLM as our pre-training objective for all examined models; MLM can only be expected to produce high-quality token-level representations, but not high-quality segment-level or document-level representations. We considered plausible alternatives that can address this limitation for segment-level or document-level representations that rely on Siamese Networks, such as SimCLR \cite{chen2020simple} and VICReg \cite{bardes2022vicreg}, but we do not have the resources to perform such compute-intensive experiments.

Similarly, we do not examine models on document-to-document retrieval tasks~\cite{liu-etal-2020-beyond, chalkidis-etal-2021-regulatory}, since task-specific architectures rely on Siamese Networks, i.e., encode two or three documents at a time, nor generative tasks, i.e., using Transformer-based encoder-decoder architectures, e.g., long document summarization~\cite{shen2022multilexsum}.

On another note, the scaling laws of neural language models suggest that larger and more intensively trained, i.e., trained over more data for a longer period, models perform better compared to smaller ones \cite{kaplan-2020-scaling,hoffman-chincilla-2022}. In our study, we consider models up to 150M parameters that may be considered small for today's standards, where models with billions of parameters are released; since we are compute-bound with access to limited compute resources.

Lastly, we follow a bottom-up approach, where we initially consider several alternative miniature HAT models (Section~\ref{sec:minilms}), and continue our experiments considering the most promising HAT based on the initial results to build and evaluate larger models (Section~\ref{sec:large_lms}). This approach was inevitable with respect to computational considerations. Ideally, we would like to build and evaluate larger version of all HAT variants to have a complete understanding of how different variants perform in the larger configuration.

\section*{Acknowledgments}
This work is also partly funded by the Innovation Fund Denmark (IFD)\footnote{\url{https://innovationsfonden.dk/en}} under File No.\ 0175-00011A.
This project was also supported by the TensorFlow Research Cloud (TFRC)\footnote{\url{https://sites.research.google/trc/about/}} program that provided instances of Google Cloud TPU v3-8 for free that were used to pre-train all HAT language models.

\bibliography{anthology,custom}
\bibliographystyle{acl_natbib}

\appendix

\section{Experimental Details}
\label{sec:experimental_details}

\subsection{Task-specific Architectures}
\label{sec:task-specific-archs}
We consider the following architectures per task:\vspace{2mm}

\noindent \textbf{Token Classification/Regression}: For token-level tasks, e.g., the Masked Language Modelling task described in Section~\ref{sec:evaluation_tasks}, we feed each document to HAT to produce contextualized token-level representations ($HAT^w_i$), and then feed those to a shared fully-connected projection layer (PR) to produce the final token representations ($O^w_i$):
\begin{equation}
    O^w_i = \mathrm{PR}(HAT^w_i))
    \label{eq:projection_word}
\end{equation}
\noindent PR consist of a feed-forward layer ($\mathbf{R}^H\!\xrightarrow{}\!\mathbf{R}^H$), where $H$ is model's hidden dimensionality, followed by a $\mathbf{Tanh}$ activation, similar to the one used in BERT \cite{devlin_bert_2019}.\vspace{2mm}

\noindent \textbf{Segment Classification/Regression}: For segment-level tasks, e.g., the Segment Order Prediction task described in Section~\ref{sec:evaluation_tasks}, we feed each document to HAT followed by a shared PR (applied per segment output, $\mathrm{HAT}^s_i$) to produce the final segment representations ($O^s_i$):
\begin{equation}
    O^s_i = \mathrm{PR}(\mathrm{HAT}^s_i))
    \label{eq:projection_segment}
\end{equation}
\noindent \textbf{Document Classification/Regression}: For document-level tasks, e.g., the Document Topic Classification task described in Section~\ref{sec:evaluation_tasks}, we use a max-pooling operator on top of the projected segment representations ($O^s_i$, Equation~\ref{eq:projection_segment}) to gather information across segments, followed by PR:
\begin{equation}
    D = \mathrm{PR}(\mathrm{MaxPool}([C_1, C2, \dots, C_N]))
\end{equation}

We opt the $\mathbf{MaxPool}$ operator over other alternatives ($\mathbf{MeanPool}$, $\mathbf{AttentivePool}$) based on the findings in the literature \cite{wu-etal-2021-hi, ernie-sparse}.\vspace{2mm}

\noindent \textbf{Document NLI}: For document NLI, e.g., ContractNLI described in Section~\ref{sec:evaluation_tasks}, we feed the model with the sequence of document segments and the hypothesis, i.e., each sample is formatted as $[C_1, C2, \dots, C_N, C_h]$, where $C_h$ is the hypothesis segment. We consider the last segment (output) representation as the pair (<document, hypothesis>) representation, since the last segment is the one representing the hypothesis ($C_h$), which is under examination.\vspace{2mm}

\noindent Across all four task types a classification layer ($\mathbf{R}^H\!\xrightarrow{}\!\mathbf{R}^L$), where $L$ is the number of labels, is placed on top of the final output (token, segment, document) representations to produce logits.\vspace{2mm}

\noindent \textbf{Multiple-Choice QA}: For multiple-choice QA tasks, e.g., the Multiple-Choice Masked Segment Prediction (MC-MSP) task described in Section~\ref{sec:evaluation_tasks}, we feed the model with the sequence of document segments, the query (question), if any, and one out of $K$ alternative choices at a time, i.e., each sample is formatted as $[C_1, C2, \dots, C_n, C_{Q[k]}, C_{AC[k]}]$, where $C_{Q[k]}$ is the query (question) segment, if any, and $C_{AC[k]}$ is the the $K_{th}$ alternative choice appended as a final segment. We consider the last segment (output) representation as the pair (<document, query, $K_{th}$ choice>) representation, which is fed to a fully-connected projection layer ($\mathbf{R}^H\!\xrightarrow{}\!\mathbf{R}^1$). The final model output is the sequence of all pairs scores (logits), i.e., $O=[O_1, O_2, \dots, O_K]$. Similar to NLI, we opt the last segment representation, which represents the examined choice.

\subsection{Datasets}
\label{sec:datasets}

\noindent\textbf{Midstream Tasks:} For Sentence Order Prediction (SOP) and Multiple-Choice Masked Segment Prediction (MC-MSP), we use documents from Wikipedia or C4, respectively. For the Document Topic Classification (DTC) task, we use the English part of MultiEURLEX~\cite{chalkidis-etal-2021-multieurlex} with the 20-labels set, which includes generic concepts (e.g., finance, agriculture, trade, education, etc.).\vspace{2mm}

\noindent\textbf{Donwstream Tasks:} In Table~\ref{tab:datasets}, we present details on the datasets used for downstream tasks, which where described in Section~\ref{sec:evaluation_tasks}. We use a custom split for MIMIC-III, since we consider the task of classifying discharge summaries for the 1st level concepts of the MeSH taxonomy We do so by backtracking all last-level (leaf node) original labeling to the respective 1st level concepts. This a lenient version of the original task with thousands of classes. For QuALITY, we use the standard training set of \cite{pang-2022-quality}, and split the original development subset 50/50 in two parts (custom development, and test subsets), since the labeling of the original test set is hidden, i.e., an online submission is needed to retrieve scores.

\begin{table}[h]
    \centering
    \resizebox{\columnwidth}{!}{
    \begin{tabular}{l|ccc}
        \toprule
         \bf Dataset Name  & \bf Ad-Hoc HAT & \bf HAT & \bf Longformer \\
         \midrule
         MIMIC-III & 1e-5 & 1e-5 & 1e-5 \\
         ECtHR-LJP & 1e-5 & 2e-5 & 1e-5 \\
         ContractNLI & 1e-5 & 1e-5 & 1e-5 \\
         QuALITY & 1e-5 & 2e-5 & 1e-5 \\
         ECtHR-ARG & 1e-5 & 3e-5 & 1e-5 \\
         \bottomrule
    \end{tabular}
    }
    \vspace{-2mm}
    \caption{Best learning rate used per model and task based on the performance on the development subset.}
    \label{tab:lrs}
\end{table}

\begin{figure*}[t]
    \centering
    \includegraphics[width=\textwidth]{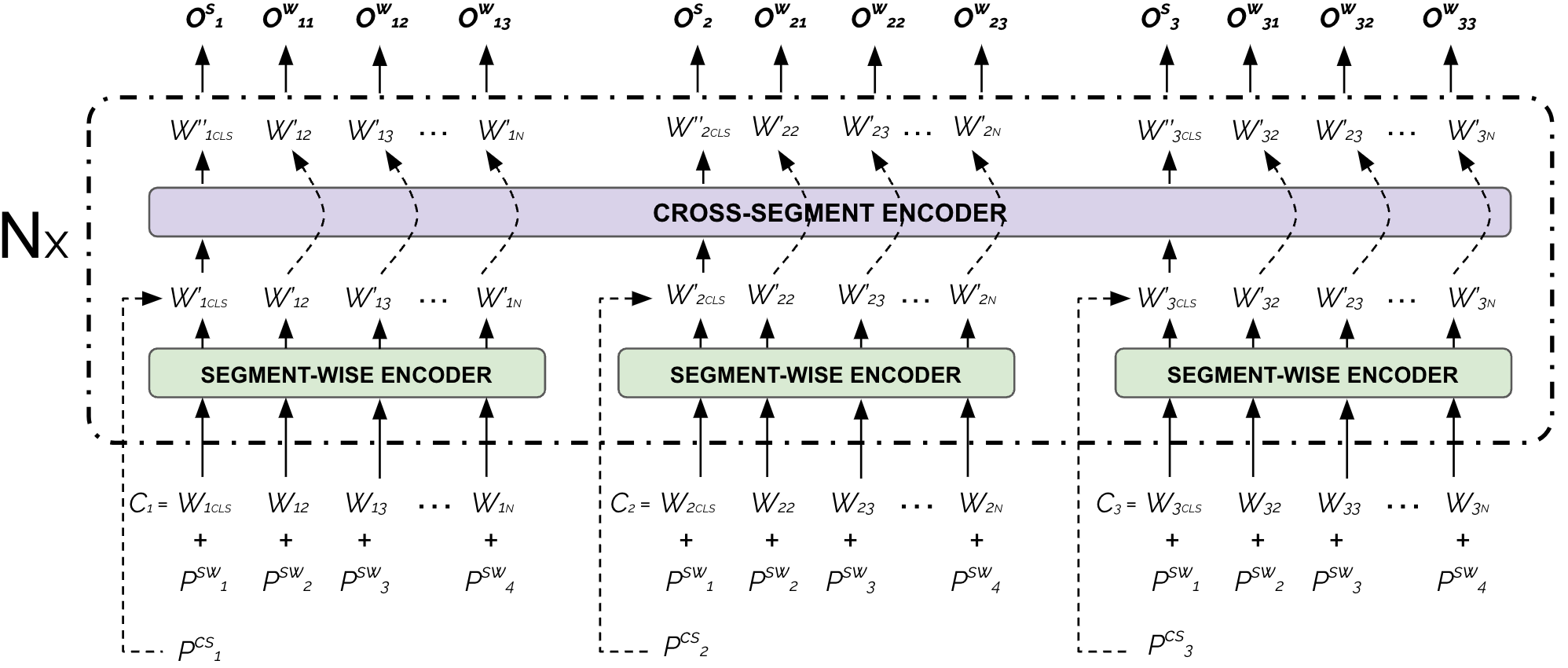}
    \caption{Example of a Hierarchical Attention Network with $N\times$ interleaved blocks.}
    \label{fig:model_example}
    \vspace{-3mm}
\end{figure*}

\subsection{Hyper-parameters}

For MLM, we use a learning rate of 1e-4, with 5\% warm-up ratio with a linear scheduling, i.e., the learning rate linearly raises up to its maximum value (1e-4) in the first 5\% of the training steps and then linearly decreases for the rest. For the rest of the tasks (midstream and downstream), we manually tune the learning-rate in \{1e-5, 2e-5\} based on performance on the development subset of each task, while we also use a 5\% warm-up ratio. We also use early stopping considering the performance on the development subset. In Table~\ref{tab:lrs}, we report the learning rates used per model and task.

\begin{table}[ht]
    \centering
    \resizebox{\columnwidth}{!}{
    \begin{tabular}{l|c|c}
         \toprule
         \bf WU Strategy & \bf Train MLM & \bf Dev MLM \\
         \midrule
         S0 & 3.10 & 2.92 \\
         S1 & 2.46 & 2.25 \\
         S2.1  & 2.35 & 2.18 \\
         S2.2 & \bf 2.34 & \bf 2.17 \\
         S2.3 & 2.46 & 2.25 \\
         \bottomrule
    \end{tabular}
    }
    \caption{MLM results (measuring cross-entropy loss) for alternative warm-up strategies (Section~\ref{sec:warmup}).}
    \label{tab:warmup}
\end{table}

\subsection{Warm-Starting}
\label{sec:warmup}

In preliminary experiments, we considered alternative warm-up strategies, i.e., initialize HAT model weights from an already pre-trained BERT (or RoBERTa) model.

\begin{itemize}[leftmargin=8pt]
\setlength\itemsep{0em}

\item\emph{(S0) None:} The first option is to not warm-up the model, and initialize all weights randomly.

\item\emph{(S1) Embeddings Only:}  The second option is to warm-up only the (word and position) embedding layers, and let all Transformer blocks randomly intiialized.

\item\emph{(S2.1) Embeddings + SW encoders:} The third option is to warm-up the embedding layers, and all segment-wise encoders, since they perform the exact same operations as the pre-trained model.

\item\emph{(S2.2) Embeddings + all encoders (Paired):} The fourth option is to warm-up the embedding layers, and all segment-wise and cross-segment encoders \emph{in pairs}, i.e., whenever a segment-wise encoder is followed by a cross-segment one, they are initialized with the very same weights.
\item\emph{(S2.3) Embeddings + all encoders (Unpaired):}  The last options is to warm-up the embedding layers, and all segment-wise and cross-segment encoders \emph{independently}, i.e., the weights of each Transformer block of the original pre-trained model are assigned to a segment-wise or cross-segment encoder.

\end{itemize}

In Table~\ref{tab:warmup}, we present the results for all alternative warm-up strategies applied to HAT (I1). We observe that any form of warm-up is better than no warm-up at all. Considering the rest of the options, the Paired warm-up leads to better MLM results.

\begin{figure*}[t]
    \centering
    \includegraphics[width=0.8\textwidth]{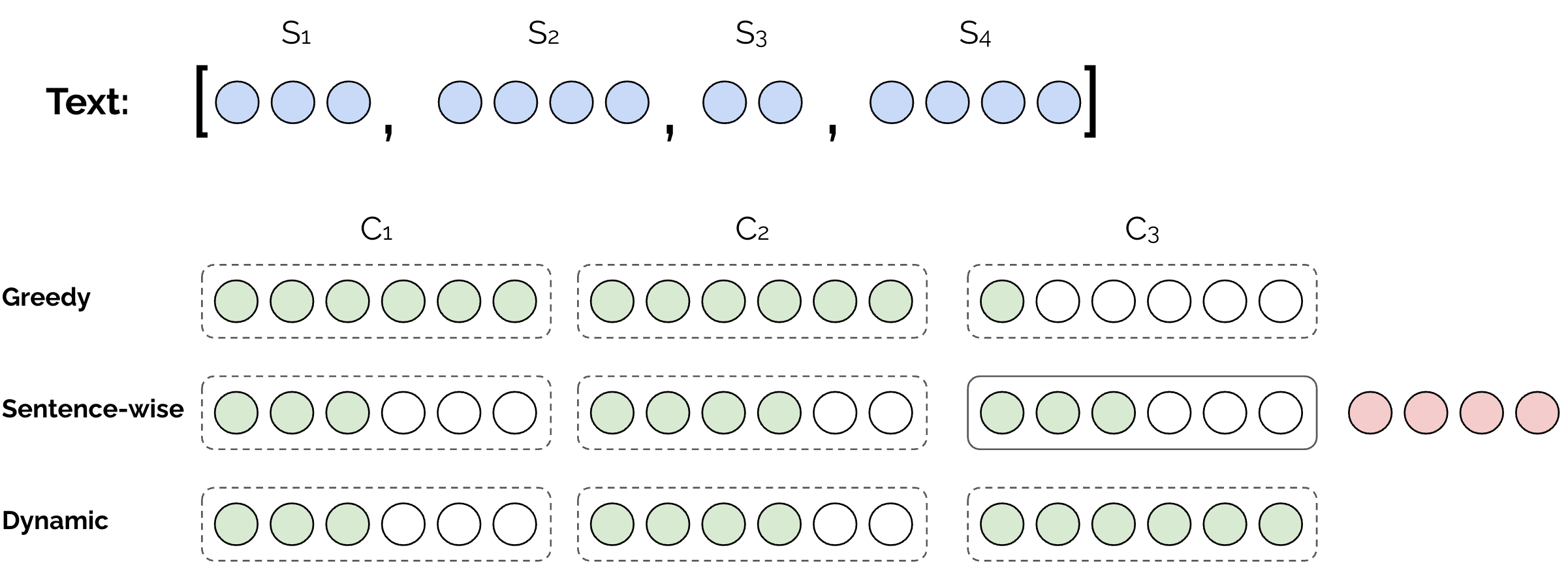}
    \caption{Text Segmentation Strategies (\emph{Greedy}, \emph{Sentence-wise}, \emph{Dynamic}). In the presented example we have a text which comprises 4 sentences, each one with a different number of tokens. Greedy segmentation leads to split sentences across segments, e.g., the last token of $S_2$ and $S_4$ has been placed in a different segment compared to the rest of the tokens. Sentence-wise segmentation leads to excessive padding and document truncation, e.g., the last sentence ($S_4$) does not fit in the models since the model can encode up to 3 segments. Dynamic segmentation avoids spliting sentences and balances padding and truncation.}
    \label{fig:segmentation_strategies}
\end{figure*}

\section{HAT Implementation Details}
\label{sec:model}

\subsection{Use of positional embeddings}
\label{sec:pos_embeds}
In Figure~\ref{fig:model_example}, we present a detailed depiction of HAT inputs and one pair of segment-wise and cross-segment encoders. The models takes as inputs sequence of tokens, organized in equally-sized segments ($[C_1, C_2, C_3]$). Special \texttt{CLS} tokens are prepended per segment. The tokens are represented by their word embedding ($W_{ij}$), and segment-wise position embedding (($P^{sw}_{j}$)). Each segment is encoded indepedently through a shared segment-wise encoder (SWE), which produced locally contextualized token representations ($W'_{ij}$). Segment representations ($W'_{i\texttt{CLS}}$) augmented with cross-segment positional embeddings (($P^{cs}_{j}$)) are fed to the successive cross-segment encoder (CSE), if any. The outputs of each block are the contextualized segment representations ($W''_{i\texttt{CLS}}$)) produced by the CS encoder, and the contextualized token representations (($W'_{ij}$)) produced by the SW encoder. HATs usually consist of a stack of such paired (or not) blocks according to the specific layout in use (as presented in Section~\ref{sec:layouts}).

\subsection{Document Segmentation Strategies}

As described in Section~\ref{sec:hier-architecture}, we opted for a \emph{dynamic} segmentation strategy. In Figure~\ref{fig:segmentation_strategies}, we present an example of the three plausible  alternatives that we considered, to express the limitation of the rest compared to the one (dynamic) used in work.

\begin{itemize}[leftmargin=8pt]
\setlength\itemsep{0em}
    \item \emph{Greedy:} In this segmentation strategy the text is split greedily in segments, i.e., there is no preservation of text structure by any means. The specific strategy is used by \citet{ernie-sparse}. While this strategy optimize for minimizing the need for truncation, it has two important limitations: (a) ignores the text structure (hierarchy), thus sentences are split at random to fill in segments, which can be proved catastrophic is specific scenarios (corner cases), where contextualization is particularly important, and (b) cannot be used for segment-level tasks.
    \item \emph{Sentence-wise:} In this segmentation strategy the text is split into sentences, i.e., each segment is equivalent to a single sentence. In this case text structure (hierarchy) is respected, but there is one crucial limitation. In case, there are lots of small-sized sentences, more than the maximum number of segments, the text will be severely truncated. In other words, many segments will be over-padded, while sentences will be truncated (not considered by the model).
    \item \emph{Dynamic:} In this segmentation strategy, the text is split into sentences, which are then grouped in larger segments up to the maximum segment length ($N$). In this case, we balance the trade-off between the preservation of the text structure (avoid sentence truncation), and the minimization of padding, which minimizes document truncation as a result. The only limitation is that sentence grouping is ad-hoc and differs across documents, since a more informed decision for sentence grouping per case (document) cannot be inferred.
\end{itemize}

\section{Computational Considerations}
\label{sec:compstats}

In order to assess the computational complexity in terms of speed (time) and memory, we conduct a controlled study, where we benchmark HAT, our Longformer,  Longformer of \citet{beltagy_longformer_2020}, and BigBird of \citet{zaheer2020bigbird} across different tasks. To account for any computational instability (hardware latency), we repeat benchmarking three times in a single NVIDIA A100 and report the best (lower) scores.\footnote{The results across runs are stable with a few exceptions; that's why we report the best to exclude outliers.} Across the three runs, we compute the averaged Batch/Sec rate, and the maximum GPU utilization (memory peak) across 100 steps.  

In the top part of Table~\ref{tab:gpumem}, we present  the Batch/Sec rate (SpeedUp) of both models, while in thw bottom part of the same table, we present the maximum GPU memory allocation. We present both measures on training (forward-backward pass) and inference (forward only) time.

\clearpage
\begin{landscape}
\centering
\begin{table}[t]
    \centering
    \resizebox{1.5\textwidth}{!}{
    \begin{tabular}{l|cc|cc|cc|cc|cc|cc|cc|cc}
         \toprule
          \bf Model Type & \multicolumn{4}{c|}{\bf Masked Language Modeling} & \multicolumn{4}{c|}{\bf Document Classification} & \multicolumn{4}{c|}{\bf Segment Classification} & \multicolumn{4}{c}{\bf Multiple-Choice QA}   \\
         \midrule
          \multicolumn{17}{c}{\bf SpeedUp (Batch/Sec)} \\
          \midrule
          & \multicolumn{2}{c}{train} & \multicolumn{2}{c|}{infer.}  & \multicolumn{2}{c}{train} & \multicolumn{2}{c|}{infer.}  & \multicolumn{2}{c}{train} & \multicolumn{2}{c}{infer.}  & \multicolumn{2}{c}{train} & \multicolumn{2}{c}{infer.}  \\
          \midrule
          Longformer (ours) & 0.266  & diff. & 0.065  & diff. & 0.210 & diff. & 0.053 & diff. & 0.459  & diff. & 0.131 & diff. & 0.386 & diff. & 0.100 & diff. \\
          HAT (ours) & 0.162 & (+39\%) & 0.051 & (+22\%) & 0.121  & (+43\%) & 0.039  & (22\%) & 0.343 & (+25\%) & 0.115 & (+14\%) & 0.207  & (+46\%) & 0.072  & (+28\%) \\
          \midrule
          Longformer (\citeyear{beltagy_longformer_2020})  &  \multicolumn{4}{c|}{---} &  0.852 & (-305\%) & 0.223 & (-321\%) & 0.895 & (-87\%) & 0.236 &  (-105\%)  &  \multicolumn{4}{c}{---} \\
          BigBird (\citeyear{zaheer2020bigbird})  &  \multicolumn{4}{c|}{---} & 0.795 & (-276\%)& 0.207 & (-291\%) & 0.795 & (-73\%) & 0.207 & (-80\%) & \multicolumn{4}{c}{---} \\
          \midrule
          \multicolumn{17}{c}{\bf  GPU Utilization} \\
          \midrule
          & \multicolumn{2}{c}{train} & \multicolumn{2}{c|}{infer.}  & \multicolumn{2}{c}{train} & \multicolumn{2}{c|}{infer.}  & \multicolumn{2}{c}{train} & \multicolumn{2}{c}{infer.}  & \multicolumn{2}{c}{train} & \multicolumn{2}{c}{infer.}  \\
          \midrule
          Longformer (ours) & 17.3GB  & diff. & 3.9GB  & diff. & 10.7GB & diff. & 1.0GB & diff . & 10.8GB & diff. & 1.0GB & diff.  & 19.3GB & diff. & 1.4GB & diff. \\
          HAT & 15.5GB (ours) & (+10\%) & 3.9GB & (0\%) & 8.9GB  & (+17\%) & 0.9GB  & (+10\%) & 8.9GB & (+18\%) & 0.9GB & (10\%) & 15.4GB  & (+20\%) & 1.2GB  & (+14\%) \\
          \midrule
          Longformer (\citeyear{beltagy_longformer_2020})   &  \multicolumn{4}{c|}{---} & 17.8GB & (-66\%) & 1.7GB & (-70\%) &  18.4GB & (-70\%) & 1.7GB & (-70\%) & \multicolumn{4}{c}{---} \\
          BigBird (\citeyear{zaheer2020bigbird})   &  \multicolumn{4}{c|}{---} & 18.8GB & (-76\%) & 1.8GB & (-80\%) &  18.9GB & (-75\%) & 1.8GB & (-80\%)  &  \multicolumn{4}{c}{---} \\
          \bottomrule
    \end{tabular}
    }
    \caption{SpeedUp (Batch/Sec) and GPU memory allocation per RoBERTa-based model (HAT, Longformer, Longformer \textregistered, and BigBird \textregistered) on NVIDIA A100.} 
    \label{tab:gpumem}
\end{table}
\end{landscape}

\end{document}